\documentclass[11pt,a4paper]{article}
\usepackage[hyperref]{emnlp2020}
\usepackage{times}
\usepackage{latexsym}

\usepackage{graphicx}
\usepackage{graphics, amsmath}
\usepackage{amsfonts}
\usepackage{amsthm}
\usepackage{amssymb}
\usepackage{multirow}
\usepackage{enumitem}
\usepackage{makecell}
\usepackage{fixltx2e}
\usepackage{bbm,color}
\usepackage{verbatim}
\usepackage{bm}
\usepackage{url}
\usepackage{microtype}
\usepackage{xcolor}
\usepackage{soul}
\usepackage{pgffor}

\aclfinalcopy 

\def\shownotes{1}  
\ifnum\shownotes=1
\newcommand{\authnote}[2]{[#1: #2]}
\else
\newcommand{\authnote}[2]{}
\fi

\title{Entity and Evidence Guided Relation Extraction for DocRED}

\author{Kevin Huang$^{\dagger}$, Guangtao Wang$^{\dagger}$, Tengyu Ma$^{\ddagger}$, Jing Huang$^{\dagger}$ \\
JD AI Research$^{\dagger}$, Stanford University$^{\ddagger}$ \\
\texttt{\{kevin.huang3, guangtao.wang, jing.huang\}@jd.com}\\ \texttt{tengyuma@stanford.edu}} 
  
\date{}

\begin{document}
\maketitle
\begin{abstract}

Document-level relation extraction is a challenging task which requires reasoning over multiple sentences in order to predict relations in a document. 
In this paper, we propose a joint training framework {\em E2GRE} (Entity and Evidence Guided Relation Extraction) for this task.
First, we introduce entity-guided sequences as inputs to a pretrained language model (e.g. BERT, RoBERTa).
These entity-guided sequences help a pretrained language model (LM) to focus on areas of the document related to the entity.
Secondly, we guide the fine-tuning of the pretrained language model by using its internal attention probabilities as additional features for evidence prediction.
Our new approach encourages the pretrained language model to focus on the entities and supporting/evidence sentences. 
We evaluate our {\em E2GRE} approach on DocRED, a recently released large-scale dataset for relation extraction. 
Our approach is able to achieve state-of-the-art results on the public leaderboard across all metrics, showing that our {\em E2GRE} is both effective and synergistic on relation extraction and evidence prediction.

\end{abstract}

\begin{figure}[h]
\framebox{
\parbox{0.45\textwidth}{
\small
\textbf{\textcolor{blue}{Relation Example} } \newline
\textbf{Document:}  {\textcolor{brown}{[0]}} \textbf{\textcolor{red}{The Legend of Zelda}} : The Minish Cap ( ) is an action - adventure game and the twelfth entry in \textbf{\textcolor{red}{The Legend of Zelda}} series.
{\textcolor{brown}{[1]}} Developed by Capcom and Flagship , with Nintendo overseeing the development process , it was released for the Game Boy Advance handheld game console in Japan and Europe in 2004 and in North America and Australia the following year .
{\textcolor{brown}{[2]}} In June 2014 , it was made available on the Wii U Virtual Console .
{\textcolor{brown}{[3]}} The Minish Cap is the third Zelda game that involves the legend of the Four Sword , expanding on the story of and .
{\textcolor{brown}{[4]}} A magical talking cap named Ezlo can shrink series protagonist \textbf{\textcolor{blue}{Link}} to the size of the Minish , a bug - sized race that live in Hyrule .
{\textcolor{brown}{[5]}} The game retains some common elements from previous Zelda installments , such as the presence of Gorons , while introducing Kinstones and other new gameplay features .
{\textcolor{brown}{[6]}} The Minish Cap was generally well received among critics .
{\textcolor{brown}{[7]}} It was named the 20th best Game Boy Advance game in an IGN feature , and was selected as the 2005 Game Boy Advance Game of the Year by GameSpot .\newline
\textbf{Head Entity:} \textbf{\textcolor{red}{The Legend of Zelda}} \newline
\textbf{Tail Entity:} \textbf{\textcolor{blue}{Link}} \newline
\textbf{Relation:} ``Publisher'' \newline
\textbf{Evidence Sentences:} 0,3,4}
}
\caption{An exemplar document in DocRED datasets where a head and tail entity pair span across multiple sentences.}
\label{fig:exp}
\end{figure}

\section{Introduction}

Relation Extraction (RE), the problem of extracting relations between pairs of entities in plain text, has received increasing research attention in recent years~\cite{zhang2017tacred,pmlr-v101-zhao19a,guo-etal-2019-attention}. It 
has important downstream applications to many other Natural Language Processing (NLP) tasks, such as Knowledge Graph Construction \cite{trisedya-etal-2019-neural}, Information Retrieval, Question Answering \cite{yu-etal-2017-improved} and Dialogue Systems \cite{Young2018aaai}. 

The majority of existing RE datasets focus on predicting \textit{intra-sentence} relations, i.e., extracting relations between entity pairs in the same sentence. For example, SemEval-2010 Task 8~\cite{hendrickx-etal-2010-semeval}, and TACRED~\cite{zhang2017tacred} are two popular RE datasets with intra-sentence relations.
These datasets have facilitated much research progress in this area such as~\cite{wang-etal-2016-relation,Alt2019,pmlr-v101-zhao19a} on SemEval-2010 Task 8 and \cite{zhang2017tacred,guo-etal-2019-attention,baldini-soares-etal-2019-matching,spanBERT2019} on TACRED. 
However, in real world applications, the majority of relations are expressed across sentences.
Figure~\ref{fig:exp} shows an example from the DocRED dataset~\citep{yao-etal-2019-docred}, which requires reasoning over three evidence sentences to predict the relational fact that ``The Legend of Zelda", is the publisher of ``Link". 

In this paper, we focus on the \textit{document-level} relation extraction problem and design a method to facilitate document-level reasoning.
We work on the DocRED~\cite{yao-etal-2019-docred}, a recent large-scale \textit{document-level} relation extraction dataset.
This dataset is annotated with a set of named entities and relations, as well as a set of supporting/evidence sentences for each relation. 
Over 40\% of the relations in DocRED require reasoning over multiple sentences. And supporting/evidence sentences can be used to provide an auxiliary task for explainable relation extraction. 

A natural attempt to solve this problem is to fine-tune the large pretrained Language Models (LMs) (e.g., GPT~\cite{radford2019GPT}, BERT~\cite{devlin-bert}, XLnet~\cite{xlnet}, RoBERTa~\cite{anonymous2020roberta}), a paradigm that has proven to be extremely successful for many NLP tasks. 
For example, all recent papers on DocRED have used BERT as an encoder to obtain the state-of-the-art results~\cite{tang2020hin,acl2020latentreasoning}. 
However, naively adapting pretrained LMs for document-level RE faces a key issue that limits its performance. Due to the length of a given document, there are more entities pairs with meaningful relations in document-level relation extraction than in the intra-sentence relation extraction.
A pretrained LM has to simultaneously encode information regarding all pairs of entities for relation extraction.
Therefore, attention values that the pretrained LM gives over all the tokens are more uniform for document-level RE compared to intra-sentence RE. 
This problem of having more uniform attention values limits the model's ability to extract information from relevant tokens from the document, limiting the effectiveness of the pretrained LM. 

In order to mitigate this problem, we propose our novel Entity and Evidence Guided Relation Extraction ({\em E2GRE}).
For each entity in a document, we generate a new input sequence by appending the entity to the beginning of a document, and then feed it into the pretrained LM. 
Thus, for each document with $N_{e}$ entities, we generate $N_{e}$ entity-guided input sequences for training.
By introducing these new training inputs, we encourage the pretrained LM to focus on the entity that is appended to the start of the document. 
We further exploit the pretrained LM by directly using internal attention probabilities as additional features for evidence prediction. 
The joint training of relation extraction and evidence prediction helps the model locate the correct semantics that are required for relation extraction. 
Both of these ideas take advantage of pretrained LMs in order to make full use of pretrained LMs for our task. 
Our main contribution is to propose the {\em E2GRE} approach, which consists of the two main ingredients below:

\begin{enumerate}

    \item  For every document, we generate multiple new inputs to feed into a pretrained language model: we concatenate every entity with the document and feed it as an input sequence to the language model. This allows the fine-tuning of the internal representations from the pretrained LM to be guided by the entity. 
    \item We further propose to use internal BERT attention probabilities as additional features for the evidence prediction.
    This allows the fine-tuning of the internal representations from the pretrained LM to be also guided by evidence/supporting sentences. 
    
\end{enumerate}

Each of these ideas give a significant boost in performance and by combining them, we are able to achieve the state-of-the-art results on DocRED leaderboard. 

\section{Related Work}
\subsection{Relation Extraction}
Relation Extraction is a long standing problem in NLP that has garnered significant research attention. 
Early work attempts to solve this problem used statistical methods with different types of feature engineering \cite{zelenko-kernel,bunescu-mooney-2005-shortest}. 
Afterwards, neural models have shown better performance at capturing semantic relationship between entities.
These methods include CNN-based approaches \cite{zeng-etal-2014-relation,wang-etal-2016-relation} and LSTM approaches \cite{cai-etal-2016-bidirectional}.

On top of using CNNs/LSTM encoders, previous models add more layers to take advantage of these embeddings.
For example, \citet{han-etal-2018-hierarchical} introduced using hierarchical attentions in order to generate relational information from coarse-to-fine semantic ideas; \citet{zhang2017tacred} applied GCN over the pruned dependency trees, and \citet{guo-etal-2019-attention} introduced Attention Guided Graph Convolutional Networks (AG-GCNs) over dependency trees. These models have shown good performance on intra-sentence relation extraction, however, some of them are not easily adapted for inter-sentence document-level RE.


\citet{Li2016bio,quirk-poon-2017-distant,PengTACL2017} were among the early work on cross sentences and document-level relation extraction.
Most approaches for document-level RE are graph-based neural network methods.
\citet{quirk-poon-2017-distant} first introduced a document graph being used for document-level RE;
\citet{PengTACL2017} proposed a graph-structured LSTM for cross-sentence n-ary relation extraction; and \cite{song-etal-2018-n} further extended the approach to graph-state LSTM. 
In \cite{jia-etal-2019-document}, an entity-centric, multi-scale representation learning on entity/sentence/document-level LSTM model was proposed for document-level n-ary RE task. \citet{christopoulou-etal-2019-connecting} recently proposed a novel edge-oriented graph model that deviates from existing graph models.
\citet{acl2020latentreasoning} proposed an induced latent graph to perform document-level relation extraction on DocRED.
These graph models generally focus on constructing unique nodes and edges, 
and have the advantage of connecting different granularity of information and aggregate them together.


\subsection{Pretrained Language Models}
Pretrained Language Models (LMs) are powerful tools which emerged in recent years. 
Recent pretrained LMs \cite{radford2019GPT,devlin-bert,xlnet,anonymous2020roberta} are Transformer-based \cite{Vaswani17transformers}, and trained with enormous amounts of data. \cite{devlin-etal-2019-bert} was the first large pretrained transformer-based LM to be released, and immediately get the state-of-the-art performance on a number of NLP tasks. 
New pretrained LM models such as XLNet \cite{xlnet} and RoBERTa~\cite{anonymous2020roberta} further increase the performance on the most NLP tasks. 

In order to take advantage of the large amounts of text that these models have seen, we finetune all of the weights inside the model. 
Finetuning on large pretrained LMs has been shown to be effective on relation extraction~\cite{wadden-etal-2019-entity}. 
Generally, large pretrained LMs are used to encode a sequence and then generate the representation of a head/tail entity pair to learn a classification \cite{SpBERT, docred}. 
\citet{baldini-soares-etal-2019-matching} introduced a new concept similar to BERT called ``matching-the-black'' and pretrained a Transformer-like model for relation learning. 
The models were fine-tuned on SemEval-2010 Task 8 and TACRED achieved state-of-the-art results.
Our method aims to improve the effectiveness of a pretrained LMs, and directly influence the finetuning of the pretrained LMs with our entity and evidence guided approach.

\section{Methods}

In this section, we introduce our {\em E2GRE} method. We first describe how to generate entity-guided inputs in Section~\ref{subsec:inputs}. Then, we present the entity-guided RE (Relation Extraction) in Section~\ref{subsec:RE}. 
Finally, we describe the entity and evidence-guided joint training for RE in Section~\ref{subsec:evidencePred}.
We use BERT as an embodiment of a pretrained LM, and use BERT when describing our methods.
\begin{figure*}[t]
    \centering
    \includegraphics[width=1.0\linewidth]{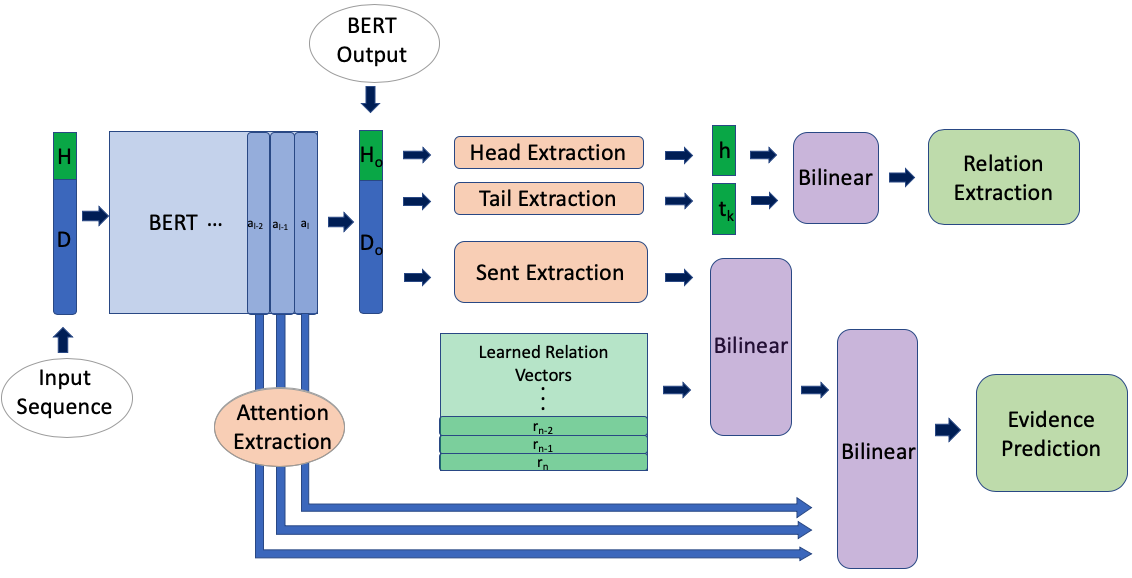}
    \caption{Diagram of our {\em E2GRE} framework. As shown in the diagram, we pass an input sequence consisting of an entity and document into BERT. We extract head and tails for relation extraction. We have learned relation vector weights shown in green. We also extract out sentence, relation vectors, and BERT attention probabilities for evidence predictions.}
    
    \label{fig:model}
\end{figure*}
\subsection{Entity-Guided Input Sequences}\label{subsec:inputs}

The relation extraction task is to predict the {\em relation} between each pair of {\em head} entity and {\em tail} entity in a given {\em document}. 

We design the entity-guided inputs to give BERT more guidance towards the entities when finetuning.
Each training input is organized by concatenating the tokens of the first mention of a single entity, denoted by $H$ (named \textit{Concatenated Head Entity}), together with the document tokens $D$, to form: ``[CLS]''+ $H$ + ``[SEP]'' + $D$ + ``[SEP]'', which is then fed into BERT. 
We generate such input sequences for each entity in the given document. 
Therefore, for a document with $N_e$ entities, $N_e$ new entity-guided input sequences are generated and fed into BERT separately.

Due to BERT's sequence length constraint of 512 tokens, if the length of the training input is longer than 512, we make use of a sliding window approach over the document:
we separate the input into multiple sequences. The first sequence is the original input sequence up to 512 tokens. The second sequence is the same as the first sequence, with an offset to the document, such that it can reach the end.
This is shown as ``[CLS]''+ $H$ + ``[SEP]'' + $D$[offset:end] + ``[SEP]''. We combine these two input sequences in our model by averaging the embeddings, and compute the BERT attention probabilities of the tokens twice in the model.

\subsection{Entity-Guided Relation Extraction}\label{subsec:RE}





For a given training input, we have one head entity, which corresponds with the concatenated entity $H$ in the input, and $N_{e}-1$ different tail entities, which are located within the document $D$.
Our method predicts $N_{e}-1$ different relations for each training input, corresponding to $N_{e}-1$ head/tail entity pairs. 

After passing a training input through BERT, we extract out the {\em head entity} embedding and a set of {\em tail entity} embeddings from the BERT output.
We average the embeddings over the concatenated head entity tokens to obtain the head entity embedding $\bm{h}$.
This is shown as the \textit{Head Extraction} in Fig. \ref{fig:model}. 
In order to extract the $k$-th tail entity embedding $\bm{t}_{k}$, we locate the indices of the tokens of $k$-th tail entity, average the output embeddings of BERT at these indices to get $\bm{t}_{k}$ (i.e., \textit{Tail Extraction} in Fig. \ref{fig:model}).

After obtaining the head entity embedding $\bm{h} \in \mathbb{R}^{d}$ and all tail entity embeddings $\{\bm{t}_{k}|\bm{t}_{k} \in \mathbb{R}^{d}$\} in a entity-guided sequence, where $1\leq k \leq N_{e}-1$, we feed them into a bilinear layer with the sigmoid activation function to predict the probability of $i$-th relation between the head entity $h$ and the $k$-th tail entity $t_{k}$, denoted by $\hat{y}_{ik}$, as follows
\begin{equation}
    \hat{y}_{ik} = \delta(\bm{h}^{T}\bm{W}_{i} \bm{t}_{k} + b_{i})
\end{equation}
where $\delta$ is the sigmoid function, $\bm{W}_i$ and $b_i$ are the learnable parameters corresponding to $i$-th relation, where $1\leq i \leq N_{r}$, and $N_r$ is the number of relations. 

Finally, we finetune BERT with a multi-label cross-entropy loss as follow:
\begin{align}
L_{RE} &= -\frac{1}{N_{r}}\frac{1}{N_{e}-1} \sum_{k=1}^{N_{e}-1} \sum_{i=1}^{N_{r}} (y_{ik}log(\hat{y}_{ik})\nonumber\\
 & +(1-y_{ik})log(1-\hat{y}_{ik}))
\end{align}

During inference, the goal of relation extraction is to predict a relation for each pair of head/tail entity within a document.
For a given entity-guided input sequence of ``[CLS]''+ entity + ``[SEP]'' + document + ``[SEP]'', the output of our model is a set of $N_{e} -1 $ relation predictions.
We combine the predictions from every sequence generated from the same document and with different head entity, in order to obtain all relation predictions over the document. 

\subsection{Evidence Guided Relation Extraction}
\label{subsec:evidencePred}

\subsubsection{Evidence Prediction}
\label{subsec:evid}


Evidence/supporting sentences are the sentences containing important supporting facts for predicting the correct relationships between head and tail entities. Therefore, evidence prediction is a good auxiliary task to relation extraction and also provides explainability for the model. 

The objective of evidence prediction is to predict whether a given sentence is evidence/supporting sentence for a given relation.
Let $N_{s}$ be the number of sentences in the document. We first obtain the sentence embedding $\bm{s} \in \mathbb{R}^{N_{S} \times d}$ by averaging all the embeddings of the words in $s$ (i.e., \textit{Sentence Extraction} in Fig. \ref{fig:model}). 
These word embeddings are derived from the BERT output embeddings. 

Let $\bm{r}_{i} \in \mathbb{R}^{d}$ be the relation embedding of $i$-th relation ($1\leq i \leq N_{r}\}$), which is initialized randomly and learnable in our model. 
We employ a bilinear layer with sigmoid activation function to predict the probability of the $j$-th sentence $s_j$ being a supporting sentence w.r.t. the given $i$-th relation $r_i$ as follows.
\begin{align}\label{eq:sentPred}
\bm{f}^{i}_{jk} &= \bm{s}_{j}\bm{W}_{i}^{r}\bm{r}_i + b_{i}^{r} \nonumber\\
    \hat{\bm{y}}^{i}_{jk} &= \delta(\bm{f}^{i}_{jk} \bm{W}_{o}^{r} + b_{o}^{r})
\end{align}

    
    
where $\bm{s}_{j}$ represents the embedding of $j$th sentence, $\bm{W}_{i}^{r}/b_{i}^{r}$ and $\bm{W}_{o}^{r}/b_{o}^{r}$ are the learnable parameters w.r.t. $i$-th relation.
We define the loss of evidence prediction under the given $i$-th relation as follows:
\begin{align}
    L_{Evi} &= -\frac{1}{N_t} \frac{1}{N_s}\sum_{k=1}^{N_t}\sum_{j=1}^{N_s} (y^{i}_{jk}log(\hat{\bm{y}}^{i}_{jk})\nonumber\\ 
    &+(1-y^{i}_{jk})log(1-\hat{\bm{y}}^{i}_{jk}))
\end{align}
where $y_{ik}^{j} \in \{0, 1\}$, and $y_{ik}^{j} = 1$  means that sentence $j$ is an evidence for inferring $i$-th relation.
It should be noted that in the training stage, we use the embedding of true relation in Eq. \ref{eq:sentPred}. In testing/inference stage, we use the embedding of the relation predicted by the relation extraction model in Section \ref{subsec:RE}.

\subsubsection{Evidence-guided Finetuning with BERT Attention Probabilities}\label{subsec:evid-bert}


Internal attention probabilities of BERT help locate the areas within a document where the BERT model focuses on. 
Therefore, these probabilities can guide the language model to focus on relevant areas of the document for relation extraction (See the attention visualization in Section \ref{sec:attn_viz}). 
In fact, we find that the areas with higher attention values are usually come from the supporting sentences. Therefore, we believe these attention probabilities can be helpful for evidence prediction. 
For each pair of head $h$ and tail $t_k$, we make use of the attention probabilities extracted from the last $l$ internal BERT layers for evidence prediction.

Let $\bm{Q} \in \mathbb{R}^{N_{h} \times L \times (d/N_{h})}$ be the query and $\bm{K} \in \mathbb{R}^{N_{h} \times L \times (d/N_{h})} $ be the key of the multi-head self attention layer, $N_{h}$ be the number of attention heads as described in \cite{Vaswani17transformers}, $L$ be the length of the input sequence (i.e., the length of entity-guided sequence defined in Section \ref{subsec:RE}) and $d$ being the embedding dimension.
We first extract the output of multi-headed self attention (MHSA) $\bm{A} \in \mathbb{R}^{N_{h}\times L \times L}$ from a given layer in BERT as follows. 
These extraction outputs are shown as ``Attention Extractor'' in Fig. \ref{fig:model}. 
\begin{align}
    & \textup{Attention} = \textup{softmax}(\frac{\bm{Q}\bm{K}^{T}}{\sqrt{d/N_{h}}}) \\
    & \textup{Att-head}_{i} = \textup{Attention}(\bm{Q}\bm{W}_{i}^{Q}, \bm{K}\bm{W}_{i}^{K})\\
    & \bm{A} =\textup{Concat}(\textup{Att-head}_{i}, \cdots ,\textup{Att-head}_{n})\label{eq:layerAttn}
\end{align}
For a given pair of head $h$ and tail $t_{k}$, we extract the attention probabilities corresponding to head and tail tokens to help relation extraction. Specifically, we concatenate the MHSAs for the last $l$ BERT layers extracted by Eq. \ref{eq:layerAttn} to form an attention probability tensor as: $\tilde{\bm{A}_{k}} \in \mathbb{R}^{l \times N_{h}\times L \times L}$.

Then, we calculate the attention probability representation of each sentence under a given head-tail entity pair as follows. 
\begin{enumerate}
    \item We first apply maximum pooling layer along the attention head dimension (i.e., second dimension) over $\tilde{\bm{A}}_{k}$. The max values are helpful to show where a specific attention head might be looking at. Afterwards we apply mean pooling over the last $l$ layers. 
    We obtain $\tilde{\bm{A}}_{s} = \frac{1}{l}\sum_{i=1}^{l}\textup{maxpool}(\tilde{\bm{A}}_{ki})$,  $\tilde{\bm{A}}_{s} \in \mathbb{R}^{L \times L}$ from these two steps.
    
    \item We then extract the attention probability tensor from the head and tail entity tokens according to the start and end positions of in the document.
    We average the attention probabilities over all the tokens for the head and tail embeddings to obtain $\tilde{\bm{A}}_{sk} \in \mathbb{R}^{L}$.

    \item Finally, we generate sentence representations from $\tilde{\bm{A}}_{sk}$ by averaging over the attentions of each token in a given sentence from the document to obtain $\bm{a}_{sk} \in \mathbb{R}^{N_{s}}$

\end{enumerate}


Once we get the attention probabilities $\bm{a}_{sk}$, we combine $\bm{a}_{sk}$ with the evidence prediction result $\hat{\bm{y}}^{s}_{ik}$ of sentence $s$ from Eq.~\ref{eq:sentPred} to form the new sentence representation and feed it into a bilinear layer with sigmoid for evidence sentence prediction as follows:
\begin{align}\label{eq:attenSentPred}
    \bm{\hat{y}}^{ia}_{k} = \delta(\bm{a}_{sk}\bm{W}_{i}^{a} \bm{f}_{k}^{i} + b_i^{a})
\end{align}
where $\bm{f}_{k}^{i}$ is the vector of fused representation of sentence embeddings and relation embeddings for a given head/tail entity pair.

Finally, we define the loss of evidence prediction under a given $i$-th relation based on attention probability representation as follows: 
\begin{align}
    L_{Evi}^{a} &= -\frac{1}{N_t} \frac{1}{N_s}\sum_{k=1}^{N_t}\sum_{j=1}^{N_s} (y^{ia}_{jk}log(\hat{\bm{y}}^{ia}_{jk})\nonumber\\ 
    &+(1-y^{ia}_{jk})log(1-\hat{\bm{y}}^{ia}_{jk}))
\end{align}
where $\hat{\bm{y}}^{ia}_{jk}$ is the $j$-th value of $\hat{\bm{y}}^{ia}_{k}$ computed by Eq. \ref{eq:attenSentPred}.

\subsubsection{Joint Training with Evidence Prediction}
We combine the relation extraction loss and attention probability guided evidence prediction loss as the final objective function for the joint training:
\begin{align}
      & Loss = L_{RE}+\lambda_{1}*L_{Evi}^{a}
\end{align}
where $\lambda_{1} > 0$ is the weight factor to make trade-offs between two losses, which is data dependent.

\section{Experiments}
We present the experimental results of our model {\em E2GRE} and compare with previously established baselines and published results, as well as the public leaderboard results on DocRED.

\subsection{Dataset}

\noindent DocRED \cite{docred} is a large document-level data set for the tasks of relation extraction and evidence sentence prediction. 
It consists of $5053$ documents, $132375$ entities, and $56354$ relations mined from Wikipedia articles. 
For each (head, tail) entity pair, there are $97$ different relation types as the candidates to predict. 
The first relation type is an ``NA'' relation between two entities, and the rest of them corresponds to a WikiData relation name.
Each of the head/tail pair that contain valid relations also include a set of supporting/evidence sentences. 

We follow the same setting in~\cite{docred} to split the data into Train/Validation/Test for model evaluation to make a fair comparison. The number of documents in Train/Validation/Test is $3000$/$1000$/$1000$, respectively.

The dataset is evaluated with the metrics of relation extraction \textbf{RE F1}, and evidence \textbf{Evi F1}.
There are also instances where relational facts may occur in the validation and train set, and so we also evaluate on the \textbf{Ign RE F1}, which removes these relational facts.

\subsection{Experimental Setup}
\noindent\textbf{hyper-parameter Setting.} The configuration for the BERT-base model follows the setting in \cite{devlin-bert}. We set the learning rate as 1e-5, $\lambda_{1}$ as 1e-4, the hidden dimension of the relation vectors as $108$, and extract internal attention probabilities from last three BERT layers. 
 
We conduct most of our experiments by fine-tuning the BERT-base model. The implementation is based on the PyTorch~\cite{paszke2017automatic} implementation of BERT\footnote{https://github.com/huggingface/pytorch-pretrained-BERT}. We run our model on a single V100 GPU for 60 epochs, resulting in approximately one day of training. The DocRED baseline and our {\em E2GRE} model have ~115M parameters\footnote{We will release the code after paper review.}.

\noindent\textbf{Baseline Methods}. We compare our model with the following published models.

\noindent1. \textit{Context Aware BiLSTM}. \citet{docred} introduced the original baseline to DocRED in their paper. They used a context-aware BiLSTM (+ additional features such as entity type, coreference and distance) to encode the document. Head and tail entities are then extracted for relation extraction.

\noindent2. \textit{BERT Two-Step}. \citet{twostepBert} introduced finetuning BERT in a two-step process, where the model first does predicts the NA relation, and then predicts the rest of the relations.\footnote{BERT Two-Step is an arxiv preprint}.

\noindent3. \textit{HIN}. \citet{tang2020hin} introduced using a hierarchical inference network to help aggregate the information from entity to sentence and further to document-level in order to obtain semantic reasoning over an entire document.

\noindent4. \textit{BERT+LSR}. \citet{acl2020latentreasoning} introduced using an induced latent graph structure to help learning how the information should flow between entities and sentences within a document.

\subsection{Main Results}
As shown in Table \ref{table:leaderboard}, our method {\em E2GRE} is the current state-of-the-art model on the public leaderboard for DocRED. 

Table~\ref{table:re} compares our method with the baseline models. From Table~\ref{table:re}, we observe that our {\em E2GRE} method is not only competitive to the previous best methods on the development set, but also holds the following advantages over previous models.

\begin{table}[!h]
\begin{small}
\begin{center}
\begin{tabular}{|l|r|r|r|}
\hline \bf User & \bf RE Ign F1 & \bf Re F1(\%) & \bf Evi F1(\%)\\ \hline
BigOrange & 60.1 & 62.3 & - \\
nttmac  & 60.2 & 63.3 & - \\
Ours  & \bf 60.3 & \bf 62.5 & \bf 50.5\\ 
\hline
\end{tabular}
\end{center}
\end{small}
\caption{\label{font-table} Top public leaderboard numbers on DocRED. Our {\em E2GRE} method uses RoBERTa-large.
}
\label{table:leaderboard}
\end{table}

\begin{table}[!h]
\begin{small}
\begin{center}
\resizebox{1.0\columnwidth}{!}{
\begin{tabular}{|l|l|l|l|}
\hline \bf Model & \bf Ign F1(\%) & \bf RE F1(\%) & \bf Evi F1 \\ \hline
\shortstack[l]{Context-Aware\\~\cite{docred}} & 48.94 & 51.09 & - \\
\hline
\shortstack[l]{BERT Two-Step\\~\cite{twostepBert}}& - & 54.42 & - \\
\hline
\shortstack[l]{HIN-BERT\\~\cite{tang2020hin}}& 54.29 & 56.31 & - \\
\hline
\shortstack[l]{BERT + LSR\\~\cite{acl2020latentreasoning}}& 52.43 & \bf 59.00 & - \\
\hline
E2GRE(Ours) &\bf 55.22 & 58.72 & \bf 47.12\\ \hline

\end{tabular}
}
\end{center}
\end{small}
\caption{\label{font-table} Results of relation extraction on the supervised setting of DocRED. Shown above are comparisons between {\em E2GRE}, and other published models on the validation set with BERT-base as the pretrained language model.
}
\label{table:re}
\end{table}

\noindent Our {\em E2GRE} method is not only competitive to the previous best methods on the development set, but also holds the following advantages over previous models.
\begin{itemize}
\item Our method is more intuitive and simpler in design compared to the HIN model and BERT+LSR model. In addition, our method provides interpretable relation extraction with supporting evidence prediction.



\item Our method is also better than all other models on the \textbf{Ign RE F1} metric.
This shows that our model does not memorize relational facts between entities, but rather examine relevant areas in the document to generate the correct relation extraction.
\end{itemize}
Compared to the original BERT baseline, our training time is slightly longer, due to the multiple new entity-guided input sequences. We examined with the idea of generating new sequences based on each head and tail entity pair, but such a method would scale quadratically with the number of entities in the document. Using our entity-guided approach strikes a balance between performance and the training time.

\begin{table}[!h]
\begin{small}
\begin{center}
\begin{tabular}{|l|l|l|l|}
\hline \bf Model & \bf Rec(\%) & \bf Prec(\%) & \bf F1(\%)\\ \hline
\bf Relation Extraction & & & \\ \hline
BERT + Joint Training & 53.33 & 55.79 & 54.54 \\
BERT-entity-guided  & 54.07 & \bf 60.43 & 57.08\\ 
+ Evidence Guided  &  \bf 59.09 &  56.95 &  \bf 58.72\\ \hline
\bf Evidence Prediction & & & \\ 
\hline
BERT + Joint Training & 43.48 & 41.54 & 42.49 \\
BERT-entity-guided  & 43.10 & \bf 49.66 & 46.15\\ 
+ Evidence Guided  &  \bf 48.26 &  49.47 &  \bf 47.14 \\ 
\hline
\end{tabular}
\end{center}
\end{small}
\caption{\label{font-table} Ablation study on the entity-guided vs evidence-guided RE. BERT+Joint Training is the BERT baseline with joint training of RE and evidence prediction. Results are evaluated on the validation set.
}
\label{table:ablation}
\end{table}

\subsection{Ablation Study}
\noindent\textbf{Analysis of Method Components}
Table \ref{table:ablation} shows the ablation study of our method on the effectiveness of entity-guided and evidence-guided training. The baseline here is the joint training model of relation extraction and evidence prediction with BERT-base.

We see that the entity-guided BERT improves the over this baseline by $2.5\%$, and evidence-guided training further improve the method by $~1.7\%$.
This shows that both parts of our method are important to the overall {\em E2GRE} method.
Our {\em E2GRE} method not only obtains improvement on the relation extraction F1, but it also obtains significant improvement on evidence prediction compared to this baseline. 
This further shows that our evidence-guided finetuning method is effective.

\begin{table}[!h]
\begin{small}
\begin{center}
\begin{tabular}{|l|l|l|l|}
\hline \bf Model & \bf Rec(\%) & \bf Prec(\%) & \bf F1(\%)\\ \hline
\bf Relation Extraction & & & \\ \hline 
BERT-entity-guided  & 54.07 & 60.43 & 57.08\\ 
3 Layers  & 56.50 & \bf 60.13 & \bf 58.71\\ 
6 Layers  &  \bf 61.87 &  54.14 &  58.51\\ \hline
\bf Evidence Prediction & & & \\ \hline
BERT-entity-guided  & 43.10 & \bf 49.66 & 46.15\\ 
3 Layers  & 45.33 & 49.07 & \bf 47.12\\
6 Layers  & \bf 46.34 & 48.19 & 46.90\\
\hline
\end{tabular}
\end{center}
\end{small}
\caption{\label{font-table} Ablation study on different numbers of layers of attention probabilities from BERT that are used for evidence prediction. Results are evaluated on the validation set.
}
\label{table:ablation_layers}
\end{table}

\noindent\textbf{Analysis of Number of BERT Layers.} 
We also conduct experiments to analyze the impact of the number of BERT layers used for obtaining attention probability values, see the results in Table \ref{table:ablation_layers}.
From this table, we observe that using more layers is not necessarily better for relation extraction. 
One possible reason may be that the BERT model encodes more syntactic information in the middle layers \cite{Clark2019WhatDB}.

\subsection{Attention Visualizations}\label{sec:attn_viz}
\begin{figure}[t]
    \centering
    \includegraphics[width=1.0\linewidth]{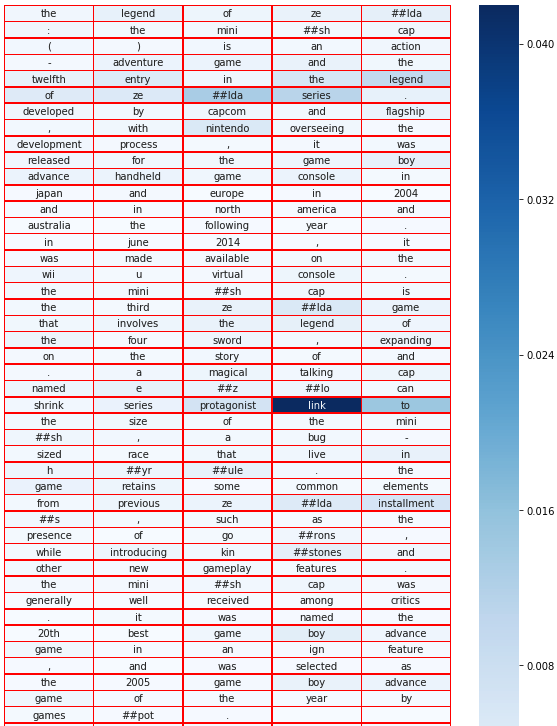}
    \caption{Baseline BERT attention heatmap over the tokenized document of a DocRED example.}
    \label{fig:baseline_attn}
\end{figure}

Fig.~\ref{fig:exp} shows an example from the validation set of our model. In this example, the relation between ``The Legend of Zelda" and ``Link" relies on information across multiple sentences in the given document. 

\begin{figure}[t]
    \centering
    \includegraphics[width=1.0\linewidth]{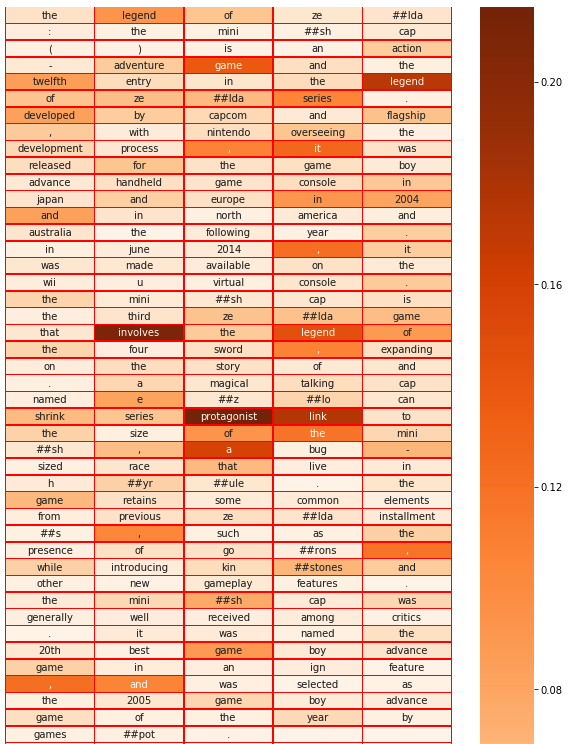}
    \caption{E2GRE's attention heatmap over the tokenized document of a DocRED example.}
    \label{fig:on_attn}
\end{figure}

Fig. \ref{fig:baseline_attn} shows the attention heatmap of naively applying BERT for relation extraction. 
This heatmap shows the attention of each word receives from `The Legend of Zelda'' and ``Link''.
We observe that the model is able to locate the relevant areas of ``Link'' and ``Legend of Zelda series'', but the attention values over the rest of the document are very small. Therefore, the model has trouble in extracting out information within the document to generate a correct relation prediction. 
 
In contrast, Fig.~\ref{fig:on_attn} shows that our {\em E2GRE} model highlights the evidence sentences, particularly in the areas where it finds relevant information. 
Phrases related to ``Link'' and ``The Legend of Zelda series'' are assigned with the higher weights. Words (such as``protagonist'' or ``involves'') linking these phrases together are also highly weighted. Moreover, the scale of the attention probabilities for E2GRE is also much larger for E2GRE compared to the baseline.
All of these phrases and bridging words are located within the evidence sentences, and make our model better at evidence prediction as well.

\section{Conclusion}
In order to more effectively exploit pretrained LMs for document-level RE, we propose a new approach called {\em E2GRE} (Entity and Evidence Guided Relation Extraction). We first generate new entity-guided sequences to feed into a LM, focusing the model on the relevant areas in the document. Then we utilize the internal attentions extracted from the last $l$ layers to help guide an LM to focus on relevant areas of the document. Our {\em E2GRE} method improves performance on both RE and evidence prediction on DocRED dataset, and achieves the state-of-the-art performance on the DocRED public leaderboard.

For future work, we plan to incorporate our ideas on using attention-guided multi-task learning to other NLP tasks with evidence sentences. Combining our approach with graph-based models for NLP tasks is another interesting direction to explore.

\bibliography{emnlp2020}                     

\begin{thebibliography}{37}
\expandafter\ifx\csname natexlab\endcsname\relax\def\natexlab#1{#1}\fi

\bibitem[{Alt et~al.(2019)Alt, H{\"{u}}bner, and Hennig}]{Alt2019}
Christoph Alt, Marc H{\"{u}}bner, and Leonhard Hennig. 2019.
\newblock \href {http://arxiv.org/abs/1906.03088} {Improving relation
  extraction by pre-trained language representations}.
\newblock \emph{CoRR}, abs/1906.03088.

\bibitem[{Baldini~Soares et~al.(2019)Baldini~Soares, FitzGerald, Ling, and
  Kwiatkowski}]{baldini-soares-etal-2019-matching}
Livio Baldini~Soares, Nicholas FitzGerald, Jeffrey Ling, and Tom Kwiatkowski.
  2019.
\newblock \href {https://doi.org/10.18653/v1/P19-1279} {Matching the blanks:
  Distributional similarity for relation learning}.
\newblock In \emph{ACL}, pages 2895--2905, Florence, Italy.

\bibitem[{Bunescu and Mooney(2005)}]{bunescu-mooney-2005-shortest}
Razvan Bunescu and Raymond Mooney. 2005.
\newblock \href {https://www.aclweb.org/anthology/H05-1091} {A shortest path
  dependency kernel for relation extraction}.
\newblock In \emph{EMNLP}, Vancouver, British Columbia, Canada.

\bibitem[{Cai et~al.(2016)Cai, Zhang, and Wang}]{cai-etal-2016-bidirectional}
Rui Cai, Xiaodong Zhang, and Houfeng Wang. 2016.
\newblock \href {https://doi.org/10.18653/v1/P16-1072} {Bidirectional recurrent
  convolutional neural network for relation classification}.
\newblock In \emph{ACL}, pages 756--765, Berlin, Germany.

\bibitem[{Christopoulou et~al.(2019)Christopoulou, Miwa, and
  Ananiadou}]{christopoulou-etal-2019-connecting}
Fenia Christopoulou, Makoto Miwa, and Sophia Ananiadou. 2019.
\newblock \href {https://www.aclweb.org/anthology/D19-1498} {Connecting the
  dots: Document-level neural relation extraction with edge-oriented graphs}.
\newblock In \emph{EMNLP}, Hong Kong, China.

\bibitem[{Clark et~al.(2019)Clark, Khandelwal, Levy, and
  Manning}]{Clark2019WhatDB}
Kevin Clark, Urvashi Khandelwal, Omer Levy, and Christopher~D. Manning. 2019.
\newblock What does bert look at? an analysis of bert's attention.
\newblock \emph{ArXiv}, abs/1906.04341.

\bibitem[{Devlin et~al.(2019{\natexlab{a}})Devlin, Chang, Lee, and
  Toutanova}]{devlin-bert}
Jacob Devlin, Ming-Wei Chang, Kenton Lee, and Kristina Toutanova.
  2019{\natexlab{a}}.
\newblock {BERT}: Pre-training of deep bidirectional transformers for language
  understanding.
\newblock In \emph{NAACL}.

\bibitem[{Devlin et~al.(2019{\natexlab{b}})Devlin, Chang, Lee, and
  Toutanova}]{devlin-etal-2019-bert}
Jacob Devlin, Ming-Wei Chang, Kenton Lee, and Kristina Toutanova.
  2019{\natexlab{b}}.
\newblock \href {https://doi.org/10.18653/v1/N19-1423} {{BERT}: Pre-training of
  deep bidirectional transformers for language understanding}.
\newblock In \emph{NAACL}, pages 4171--4186, Minneapolis, Minnesota.

\bibitem[{Eberts and Ulges(2019)}]{SpBERT}
Markus Eberts and Adrian Ulges. 2019.
\newblock Span-based joint entity and relation extraction with transformer
  pre-training.

\bibitem[{Guo et~al.(2019)Guo, Zhang, and Lu}]{guo-etal-2019-attention}
Zhijiang Guo, Yan Zhang, and Wei Lu. 2019.
\newblock \href {https://doi.org/10.18653/v1/P19-1024} {Attention guided graph
  convolutional networks for relation extraction}.
\newblock In \emph{ACL}, Florence, Italy.

\bibitem[{Han et~al.(2018)Han, Yu, Liu, Sun, and
  Li}]{han-etal-2018-hierarchical}
Xu~Han, Pengfei Yu, Zhiyuan Liu, Maosong Sun, and Peng Li. 2018.
\newblock \href {https://doi.org/10.18653/v1/D18-1247} {Hierarchical relation
  extraction with coarse-to-fine grained attention}.
\newblock In \emph{EMNLP}, Brussels, Belgium.

\bibitem[{Hendrickx et~al.(2010)Hendrickx, Kim, Kozareva, Nakov,
  {\'O}~S{\'e}aghdha, Pad{\'o}, Pennacchiotti, Romano, and
  Szpakowicz}]{hendrickx-etal-2010-semeval}
Iris Hendrickx, Su~Nam Kim, Zornitsa Kozareva, Preslav Nakov, Diarmuid
  {\'O}~S{\'e}aghdha, Sebastian Pad{\'o}, Marco Pennacchiotti, Lorenza Romano,
  and Stan Szpakowicz. 2010.
\newblock \href {https://www.aclweb.org/anthology/S10-1006} {{S}em{E}val-2010
  task 8: Multi-way classification of semantic relations between pairs of
  nominals}.
\newblock In \emph{Proceedings of the 5th International Workshop on Semantic
  Evaluation}, pages 33--38, Uppsala, Sweden. ACL.

\bibitem[{Jia et~al.(2019)Jia, Wong, and Poon}]{jia-etal-2019-document}
Robin Jia, Cliff Wong, and Hoifung Poon. 2019.
\newblock \href {https://doi.org/10.18653/v1/N19-1370} {Document-level n-ary
  relation extraction with multiscale representation learning}.
\newblock In \emph{NAACL}, Minneapolis, Minnesota.

\bibitem[{Joshi et~al.(2019)Joshi, Chen, Liu, Weld, Zettlemoyer, and
  Levy}]{spanBERT2019}
Mandar Joshi, Danqi Chen, Yinhan Liu, Daniel~S. Weld, Luke Zettlemoyer, and
  Omer Levy. 2019.
\newblock \href {http://arxiv.org/abs/1907.10529} {Spanbert: Improving
  pre-training by representing and predicting spans}.
\newblock \emph{CoRR}, abs/1907.10529.

\bibitem[{Li et~al.(2016)Li, Sun, Johnson, Sciaky, Wei, Leaman, Davis,
  Mattingly, Wiegers, and Lu}]{Li2016bio}
Jiao Li, Yueping Sun, Robin~J. Johnson, Daniela Sciaky, Chih-Hsuan Wei, Robert
  Leaman, Allan~Peter Davis, Carolyn~J. Mattingly, Thomas~C. Wiegers, and
  Zhiyong Lu. 2016.
\newblock {BioCreative V CDR task corpus: a resource for chemical disease
  relation extraction}.
\newblock \emph{Database}, 2016.

\bibitem[{Nan et~al.(2020)Nan, Guo, Sekulić, and Lu}]{acl2020latentreasoning}
Guoshun Nan, Zhijiang Guo, Ivan Sekulić, and Wei Lu. 2020.
\newblock Reasoning with latent structure refinement for document-level
  relation extraction.
\newblock In \emph{ACL}.

\bibitem[{Paszke et~al.(2017)Paszke, Gross, Chintala, Chanan, Yang, DeVito,
  Lin, Desmaison, Antiga, and Lerer}]{paszke2017automatic}
Adam Paszke, Sam Gross, Soumith Chintala, Gregory Chanan, Edward Yang, Zachary
  DeVito, Zeming Lin, Alban Desmaison, Luca Antiga, and Adam Lerer. 2017.
\newblock Automatic differentiation in pytorch.

\bibitem[{Peng et~al.(2017)Peng, Poon, Quirk, Toutanova, and tau
  Yih}]{PengTACL2017}
Nanyun Peng, Hoifung Poon, Chris Quirk, Kristina Toutanova, and Wen tau Yih.
  2017.
\newblock Cross-sentence n-ary relation extraction with graph lstms.
\newblock \emph{TACL}, 5(0).

\bibitem[{Quirk and Poon(2017)}]{quirk-poon-2017-distant}
Chris Quirk and Hoifung Poon. 2017.
\newblock \href {https://www.aclweb.org/anthology/E17-1110} {Distant
  supervision for relation extraction beyond the sentence boundary}.
\newblock In \emph{ACL}, pages 1171--1182, Valencia, Spain. ACL.

\bibitem[{Radford et~al.(2019)Radford, Wu, Child, Luan, Amodei, and
  Sutskever}]{radford2019GPT}
Alec Radford, Jeff Wu, Rewon Child, David Luan, Dario Amodei, and Ilya
  Sutskever. 2019.
\newblock Language models are unsupervised multitask learners.

\bibitem[{Song et~al.(2018)Song, Zhang, Wang, and Gildea}]{song-etal-2018-n}
Linfeng Song, Yue Zhang, Zhiguo Wang, and Daniel Gildea. 2018.
\newblock \href {https://doi.org/10.18653/v1/D18-1246} {N-ary relation
  extraction using graph-state {LSTM}}.
\newblock In \emph{EMNLP}, Brussels, Belgium.

\bibitem[{Tang et~al.(2020)Tang, Cao, Zhang, Cao, Fang, Wang, and
  Yin}]{tang2020hin}
Hengzhu Tang, Yanan Cao, Zhenyu Zhang, Jiangxia Cao, Fang Fang, Shi Wang, and
  Pengfei Yin. 2020.
\newblock Hin: Hierarchical inference network for document-level relation
  extraction.
\newblock In \emph{PAKDD}.

\bibitem[{Trisedya et~al.(2019)Trisedya, Weikum, Qi, and
  Zhang}]{trisedya-etal-2019-neural}
Bayu~Distiawan Trisedya, Gerhard Weikum, Jianzhong Qi, and Rui Zhang. 2019.
\newblock \href {https://doi.org/10.18653/v1/P19-1023} {Neural relation
  extraction for knowledge base enrichment}.
\newblock In \emph{ACL}, pages 229--240, Florence, Italy. ACL.

\bibitem[{Vaswani et~al.(2017)Vaswani, Shazeer, Parmar, Uszkoreit, Jones,
  Gomez, Kaiser, and Polosukhin}]{Vaswani17transformers}
Ashish Vaswani, Noam Shazeer, Niki Parmar, Jakob Uszkoreit, Llion Jones,
  Aidan~N. Gomez, Lukasz Kaiser, and Illia Polosukhin. 2017.
\newblock \href {http://arxiv.org/abs/1706.03762} {Attention is all you need}.
\newblock \emph{NeurIPS}, abs/1706.03762.

\bibitem[{Wadden et~al.(2019)Wadden, Wennberg, Luan, and
  Hajishirzi}]{wadden-etal-2019-entity}
David Wadden, Ulme Wennberg, Yi~Luan, and Hannaneh Hajishirzi. 2019.
\newblock \href {https://doi.org/10.18653/v1/D19-1585} {Entity, relation, and
  event extraction with contextualized span representations}.
\newblock In \emph{EMNLP}, Hong Kong, China.

\bibitem[{Wang et~al.(2019)Wang, Focke, Sylvester, Mishra, and
  Wang}]{twostepBert}
Hong Wang, Christfried Focke, Rob Sylvester, Nilesh Mishra, and William Wang.
  2019.
\newblock \href {http://arxiv.org/abs/1909.11898} {Fine-tune bert for docred
  with two-step process}.

\bibitem[{Wang et~al.(2016)Wang, Cao, de~Melo, and
  Liu}]{wang-etal-2016-relation}
Linlin Wang, Zhu Cao, Gerard de~Melo, and Zhiyuan Liu. 2016.
\newblock \href {https://doi.org/10.18653/v1/P16-1123} {Relation classification
  via multi-level attention {CNN}s}.
\newblock In \emph{ACL}, pages 1298--1307, Berlin, Germany. ACL.

\bibitem[{Yang et~al.(2019)Yang, Dai, Yang, Carbonell, Salakhutdinov, and
  Le}]{xlnet}
Zhilin Yang, Zihang Dai, Yiming Yang, Jaime Carbonell, Russ~R Salakhutdinov,
  and Quoc~V Le. 2019.
\newblock \href
  {http://papers.nips.cc/paper/8812-xlnet-generalized-autoregressive-pretraining-for-language-understanding.pdf}
  {Xlnet: Generalized autoregressive pretraining for language understanding}.
\newblock In \emph{Advances in Neural Information Processing Systems 32}, pages
  5754--5764. Curran Associates, Inc.

\bibitem[{Yao et~al.(2019{\natexlab{a}})Yao, Ye, Li, Han, Lin, Liu, Liu, Huang,
  Zhou, and Sun}]{yao-etal-2019-docred}
Yuan Yao, Deming Ye, Peng Li, Xu~Han, Yankai Lin, Zhenghao Liu, Zhiyuan Liu,
  Lixin Huang, Jie Zhou, and Maosong Sun. 2019{\natexlab{a}}.
\newblock \href {https://doi.org/10.18653/v1/P19-1074} {{D}oc{RED}: A
  large-scale document-level relation extraction dataset}.
\newblock In \emph{ACL}, Florence, Italy.

\bibitem[{Yao et~al.(2019{\natexlab{b}})Yao, Ye, Li, Han, Lin, Liu, Liu, Huang,
  Zhou, and Sun}]{docred}
Yuan Yao, Deming Ye, Peng Li, Xu~Han, Yankai Lin, Zhenghao Liu, Zhiyuan Liu,
  Lixin Huang, Jie Zhou, and Maosong Sun. 2019{\natexlab{b}}.
\newblock {D}oc{RED}: A large-scale document-level relation extraction dataset.
\newblock In \emph{ACL}.

\bibitem[{Yinhan~Liu(2020)}]{anonymous2020roberta}
Naman~Goyal Yinhan~Liu, Myle~Ott. 2020.
\newblock \href {https://openreview.net/forum?id=SyxS0T4tvS} {Ro{\{}bert{\}}a:
  A robustly optimized {\{}bert{\}} pretraining approach}.
\newblock In \emph{Submitted to International Conference on Learning
  Representations}.
\newblock Under review.

\bibitem[{Young et~al.(2018)Young, Cambria, Chaturvedi, Huang, Zhou, and
  Biswas}]{Young2018aaai}
Tom Young, Erik~Cambria Cambria, Iti Chaturvedi, Minlie Huang, Hao Zhou, and
  Subham Biswas. 2018.
\newblock \href
  {https://aaai.org/ocs/index.php/AAAI/AAAI18/paper/view/16573/16030}
  {Augmenting end-to-end dialog systems with commonsense knowledge}.
\newblock In \emph{AAAI}.

\bibitem[{Yu et~al.(2017)Yu, Yin, Hasan, dos Santos, Xiang, and
  Zhou}]{yu-etal-2017-improved}
Mo~Yu, Wenpeng Yin, Kazi~Saidul Hasan, Cicero dos Santos, Bing Xiang, and Bowen
  Zhou. 2017.
\newblock \href {https://doi.org/10.18653/v1/P17-1053} {Improved neural
  relation detection for knowledge base question answering}.
\newblock In \emph{ACL}, pages 571--581, Vancouver, Canada. ACL.

\bibitem[{Zelenko et~al.(2003)Zelenko, Aone, and Richardella}]{zelenko-kernel}
Dmitry Zelenko, Chinatsu Aone, and Anthony Richardella. 2003.
\newblock \href {https://doi.org/10.3115/1118693.1118703} {Kernel methods for
  relation extraction}.
\newblock \emph{Journal of Machine Learning Research}, 3:1083--1106.

\bibitem[{Zeng et~al.(2014)Zeng, Liu, Lai, Zhou, and
  Zhao}]{zeng-etal-2014-relation}
Daojian Zeng, Kang Liu, Siwei Lai, Guangyou Zhou, and Jun Zhao. 2014.
\newblock \href {https://www.aclweb.org/anthology/C14-1220} {Relation
  classification via convolutional deep neural network}.
\newblock In \emph{Proceedings of {COLING} 2014, the 25th International
  Conference on Computational Linguistics: Technical Papers}, pages 2335--2344,
  Dublin, Ireland. ACL.

\bibitem[{Zhang et~al.(2017)Zhang, Zhong, Chen, Angeli, and
  Manning}]{zhang2017tacred}
Yuhao Zhang, Victor Zhong, Danqi Chen, Gabor Angeli, and Christopher~D.
  Manning. 2017.
\newblock \href {https://nlp.stanford.edu/pubs/zhang2017tacred.pdf}
  {Position-aware attention and supervised data improve slot filling}.
\newblock In \emph{Proceedings of the 2017 Conference on Empirical Methods in
  Natural Language Processing (EMNLP 2017)}, pages 35--45.

\bibitem[{Zhao et~al.(2019)Zhao, Wan, Gao, and Lin}]{pmlr-v101-zhao19a}
Yi~Zhao, Huaiyu Wan, Jianwei Gao, and Youfang Lin. 2019.
\newblock \href {http://proceedings.mlr.press/v101/zhao19a.html} {Improving
  relation classification by entity pair graph}.
\newblock In \emph{ACML}, volume 101, Nagoya, Japan.

\end{thebibliography}
\bibliographystyle{acl_natbib}
\end{document}